\documentclass[final,3p,authoryear]{elsarticle}

\usepackage{amssymb}
\usepackage{amsmath}
\usepackage{tabularray}
\DefTblrTemplate{middlehead,lasthead}{default}{
\UseTblrTemplate{conthead}{default}
}
\usepackage{siunitx}
\usepackage{url}

\journal{Medical Image Analysis}

\begin{document}

\thispagestyle{plain}  
\vspace*{-3cm}
{\centering\copyright~2025. This manuscript version is made available under the CC-BY-NC-ND 4.0 license \url{https://creativecommons.org/licenses/by-nc-nd/4.0/}\par}
\begin{frontmatter}

\title{Application-driven Validation of Posteriors in Inverse Problems}

\author[1]{Tim J. Adler\fnref{fn1}}
\author[1,2]{Jan-Hinrich Nölke\corref{cor1}\fnref{fn1}}
\fntext[fn1]{Shared first authors}
\ead{j.noelke@dkfz-heidelberg.de}
\author[1,3]{Annika Reinke}
\author[1,4]{Minu Dietlinde Tizabi}
\author[1]{Sebastian Gruber}
\author[1]{Dasha Trofimova}
\author[5]{Lynton Ardizzone}
\author[3,6]{Paul F. Jaeger}
\author[7,8,9,10,11]{Florian Buettner}
\author[5]{Ullrich Köthe}
\author[1,2,3,4]{Lena~Maier-Hein\corref{cor1}}
\ead{l.maier-hein@dkfz-heidelberg.de}
\cortext[cor1]{Corresponding authors.}

\affiliation[1]{organization={German Cancer Research Center (DKFZ) Heidelberg, Division~of~Intelligent~Medical~Systems~(IMSY)},
            addressline={Im~Neuenheimer~Feld~280}, 
            city={Heidelberg},
            postcode={69120}, 
            country={Germany}}

\affiliation[2]{organization={Faculty of Mathematics and Computer Science, Heidelberg~University},
            addressline={Im~Neuenheimer~Feld~205}, 
            city={Heidelberg},
            postcode={69120}, 
            country={Germany}}
\affiliation[3]{organization={German Cancer Research Center (DKFZ) Heidelberg, HI~Helmholtz~Imaging},
            addressline={Im~Neuenheimer~Feld~280}, 
            city={Heidelberg},
            postcode={69120}, 
            country={Germany}}

\affiliation[4]{organization={National Center for Tumor Diseases (NCT), NCT Heidelberg, a partnership between DKFZ and University~Medical~Center~Heidelberg},
            addressline={Im~Neuenheimer~Feld~460}, 
            city={Heidelberg},
            postcode={69120}, 
            country={Germany}}

\affiliation[5]{organization={Visual Learning Lab, Interdisciplinary Center for Scientific Computing (IWR)}, 
            city={Heidelberg},
            country={Germany}}

\affiliation[6]{organization={German Cancer Research Center (DKFZ) Heidelberg, Interactive~Machine~ Learning~Group},
            addressline={Im~Neuenheimer~Feld~280}, 
            city={Heidelberg},
            postcode={69120}, 
            country={Germany}}

\affiliation[7]{organization={Department of Informatics, Goethe University Frankfurt},
            city={Frankfurt},
            country={Germany}}

\affiliation[8]{organization={Department of Medicine, Goethe University Frankfurt},
            city={Frankfurt},
            country={Germany}}

\affiliation[9]{organization={German Cancer Consortium (DKTK), partner site Frankfurt, a partnership between DKFZ and UCT~Frankfurt-Marburg},
            city={Frankfurt},
            country={Germany}}

\affiliation[10]{organization={German Cancer Research Center (DKFZ)},
            addressline={Im Neuenheimer Feld 280}, 
            city={Heidelberg},
            postcode={69120}, 
            country={Germany}}

\affiliation[11]{organization={Frankfurt Cancer Institute},
            city={Frankfurt},
            country={Germany}}

\begin{abstract}
Current deep learning-based solutions for image analysis tasks are commonly incapable of handling problems to which multiple different plausible solutions exist. In response, posterior-based methods such as conditional Diffusion Models and Invertible Neural Networks have emerged; however, their translation is hampered by a lack of research on adequate validation. In other words, the way progress is measured often does not reflect the needs of the driving practical application. Closing this gap in the literature, we present the first systematic framework for the application-driven validation of posterior-based methods in inverse problems. As a methodological novelty, it adopts key principles from the field of object detection validation, which has a long history of addressing the question of how to locate and match multiple object instances in an image. Treating modes as instances enables us to perform mode-centric validation, using well-interpretable metrics from the application perspective. We demonstrate the value of our framework through instantiations for a synthetic toy example and two medical vision use cases: pose estimation in surgery and imaging-based quantification of functional tissue parameters for diagnostics. Our framework offers key advantages over common approaches to posterior validation in all three examples and could thus revolutionize performance assessment in inverse problems.

\end{abstract}

\begin{keyword}
Validation \sep Metrics \sep Posterior \sep Deep Learning \sep Inverse Problems

\end{keyword}

\end{frontmatter}

\section{Introduction}
\label{sec:introduction}

\begin{figure*}
    \centering
    \includegraphics[width=1\textwidth]{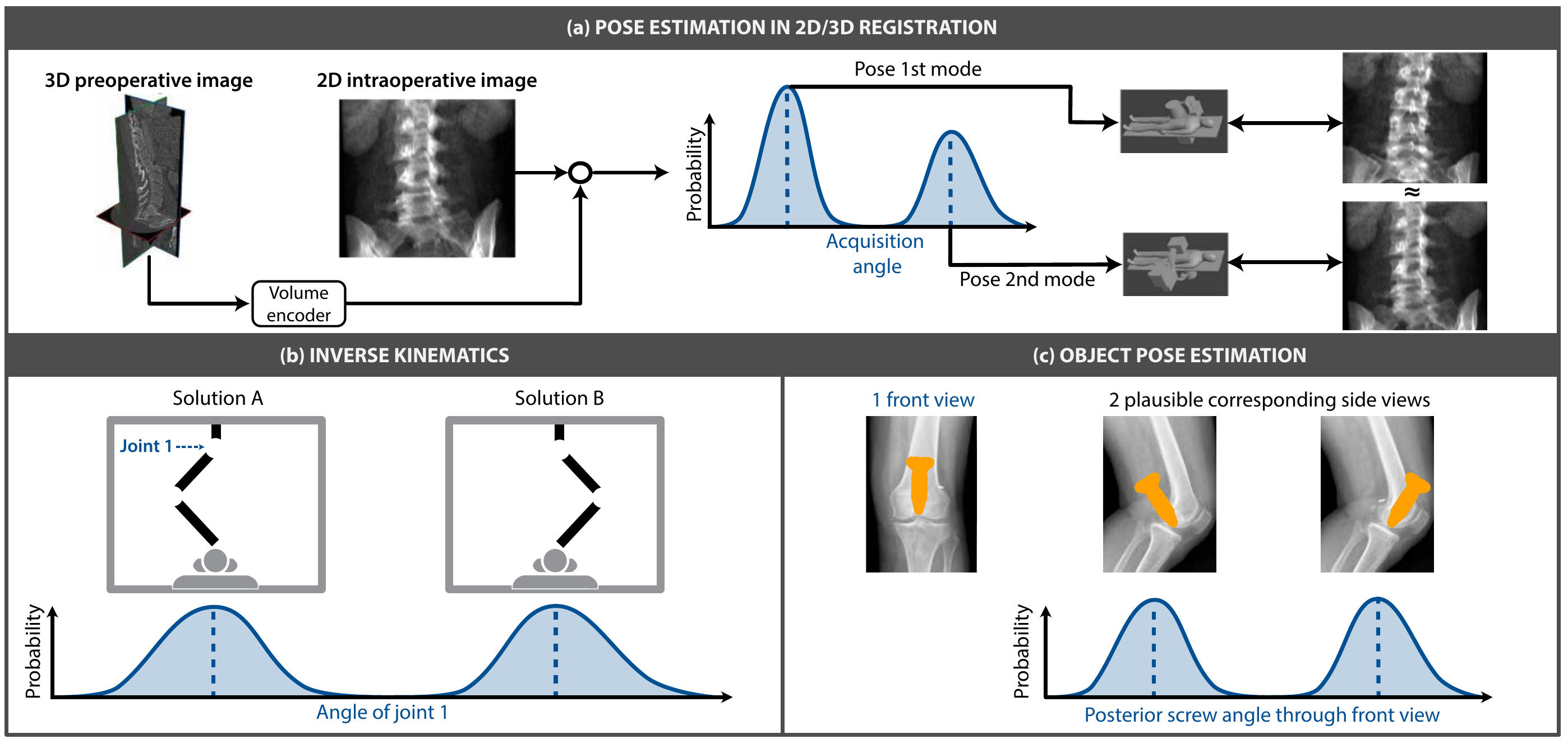}
    \caption{Computational tasks in healthcare may be inherently ambiguous. When multiple substantially different plausible solutions exist, they can be encoded as modes in the output posterior. (a): In image-guided surgery, the task of determining the pose of a patient relative to an intraoperative X-ray system may be ill-posed as different patient poses (here: posterior-anterior and anterior-posterior) can yield almost identical 2D projection images for the same device pose. (b): Inverse kinematics in robotic surgery presents inherent ambiguities when determining optimal joint configurations to reach desired target positions, as different joint angles can achieve the same end-effector position. (c):  When analyzing screw placement from a 2D frontal X-ray view, the screw angle in the lateral plane cannot be recovered due to the lack of depth information in the projection image. Knee X-ray adapted from \cite{fang_extra-articular_2021} (CC BY 4.0).
}
    \label{fig:use_cases}
\end{figure*}

\begin{figure*}
    \centering
    \includegraphics[width=1\textwidth]{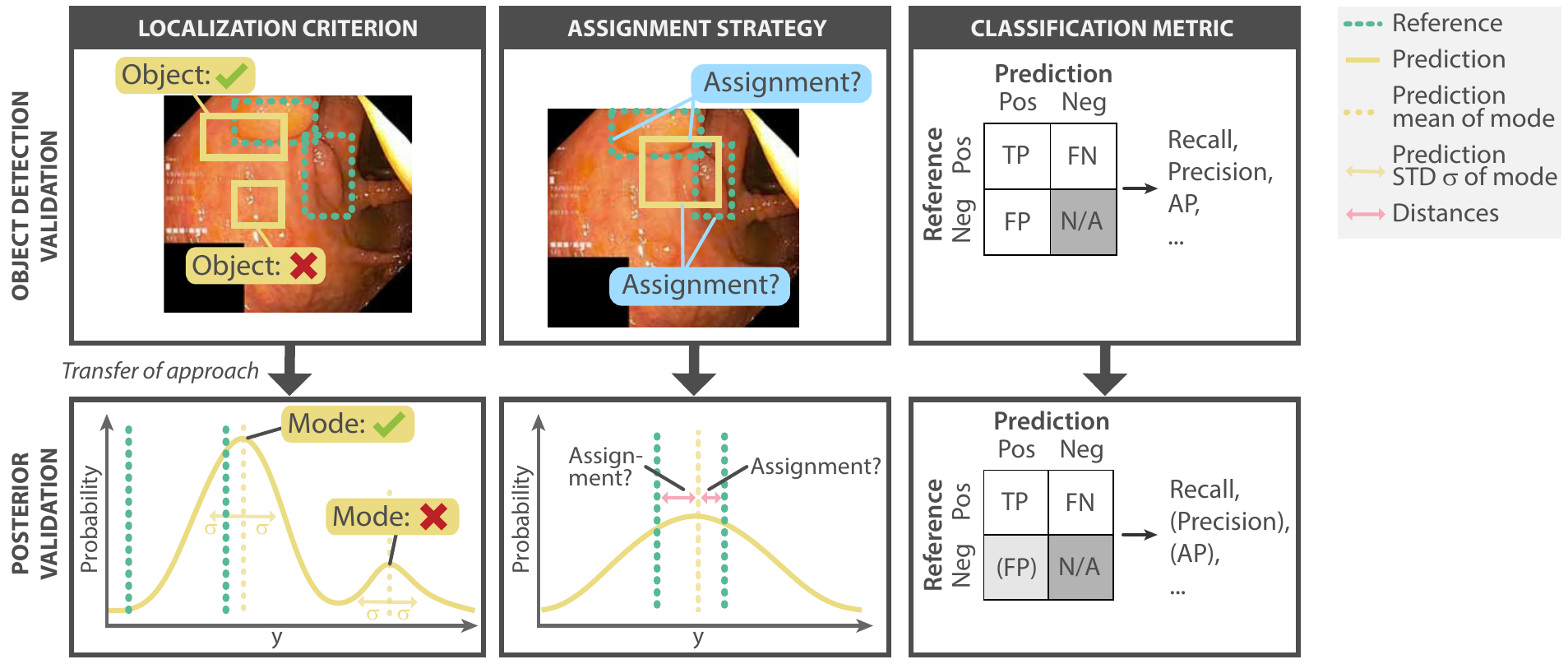}
    \caption{Object detection validation methodology lends itself well to posterior validation. This validation is subdivided into the steps of instance localization, assignment, and computing of classification metrics. These steps have natural analogs in the posterior validation case. Used abbreviations: Average Precision (AP), True Positive (TP), False Positive (FP), False Negative (FN), Standard Deviation (STD).}
    \label{fig:od_analogy}
\end{figure*}

Deep learning has led to breakthrough successes in various areas of image analysis. State-of-the-art approaches, however, commonly lack the capacity of representing the fact that multiple (substantially different) plausible solutions may exist. One medical vision example is the registration of two-dimensional (2D) X-ray images with preoperative computed tomography (CT) images in intraoperative surgical guidance systems (Fig.~\ref{fig:use_cases}). To this end, the pose of the X-ray modality relative to the patient coordinate system (given by the preoperative three-dimensional (3D) image) has to be inferred from the 2D X-ray images. Standard methods compute a single point estimate based on the input images, thereby ignoring the fact that multiple different solutions may exist. Posterior-based methods, such as conditional Diffusion Models \citep{batzolis2021conditional,chung2022improving} and Invertible Neural Networks \citep{ardizzone2018analyzing, ardizzone2019guided,nolkePhotoacousticQuantificationTissue2024}, overcome this bottleneck by converting the input to a ‘posterior’ – a full probability distribution conditioned on the input, capable of capturing several plausible solutions via multiple modes. While the field is currently experiencing much progress on a methodological level, little attention is given to the adequate validation of posterior-based methods, impeding their translation into practice.

Most commonly, methods are either validated by extracting the maximum a posteriori (MAP) probability location and using it as a point estimate \citep{ardizzone2018analyzing}, by feeding the posteriors into the forward model (if available) and choosing suitable metrics in the `observable space' \citep{ardizzone2018analyzing,zheng2022entropy,batzolis2021conditional}, or through more qualitative validation schemes as, for example, in the task of image generation \citep{ardizzone2019guided,bao2023all}. However, these validation approaches are often inadequate as they neglect the requirements imposed by the underlying application. For instance, providing the actual posteriors to the users of a system that can handle ambiguous solutions is typically not useful (lack of interpretability) and may not even be feasible (due to high dimensionality). Here, validation should rather focus on the assessment of the modes themselves, as this is what users base their decisions on in practice. In the aforementioned clinical example (Fig.~\ref{fig:use_case1_illustration}), images may commonly be acquired with the patient in supine position, i.e., lying on the back. As the prior influences the posterior, the largest mode would correspond to the standard position. The standard validation procedure based on MAP estimates would ignore the small mode corresponding to a 180° rotated pose. Ignoring the smaller mode(s) does not reflect clinical needs as a clinician could easily choose between a small set of modes and even benefit from the model information if a surprising mode appears.

In this paper, we therefore propose choosing a validation approach that reflects the requirements of the driving application. Specifically, we argue that most applications require a mode-centric validation, reflecting the fact that domain experts (e.g., clinicians) work with concrete decisions/solutions rather than probability distributions. In this vein, we propose metrics that go beyond common regression errors and directly compare (multiple) predicted to (multiple) reference modes. While this may sound trivial at first glance, the specific implementation is not straightforward (How exactly are modes localized? What to do in the case of mode assignment ambiguities?, etc.), which may be one reason why the topic has – to our knowledge – not yet been addressed in the literature. Closing this gap, our novel approach takes inspiration from the object detection community (Fig.~\ref{fig:od_analogy}). 
By adopting the principles of a localization criterion and assignment strategies, we are able to perform a mode-centric validation with much more meaningful and better interpretable metrics from an application perspective. In the above example, this approach would require defining the localization criterion (and its hyperparameters) such that the augmented reality visualization of surrounding structures in the intraoperative X-ray can be achieved with acceptable accuracy through the pose estimation. Classification-based performance metrics would then be well-interpretable by the domain expert: The False Positives Per Image (FPPI) at the computed Recall, for example, would inform the clinician that they would need to select the most plausible pose from an average of about FPPI+1 options during a surgery. While this would not be a problem for FPPI = 1, it would be infeasible to choose from, for example, ten different options.

Although it would be desirable to validate posteriors as comprehensively as possible with both distribution-based and mode-based metrics, this may not always be possible in real-world scenarios. In many cases, for example, a ground truth posterior (required for distribution-based comparison) may not be available. Moreover, the set of reference solutions may be non-exhaustive and, for example, only contain one out of possibly multiple plausible solutions. We address this challenge with a problem fingerprint that abstracts from the specific problem by capturing key problem characteristics and available data in a structured format. Guided by this fingerprint, metrics are then recommended via a decision tree.
The specific contributions of this paper are:
\newline
\begin{enumerate}
    \item Object detection analogy: To our knowledge, we are the first to uncover an analogy between validation practice in an object detection setting and validation of posteriors.
    \item Application-driven framework: Based on this analogy, we propose a posterior validation framework that takes into account both the requirements of the underlying application as well as the mathematical restrictions enforced by the available validation data. 
    \item Use case instantiation: An instantiation of the framework for three complementary use cases reveals flaws in common validation practices and showcases the benefit of a mode-centric approach.
\end{enumerate}

\begin{figure*}
    \centering
    \includegraphics[width=1\textwidth]{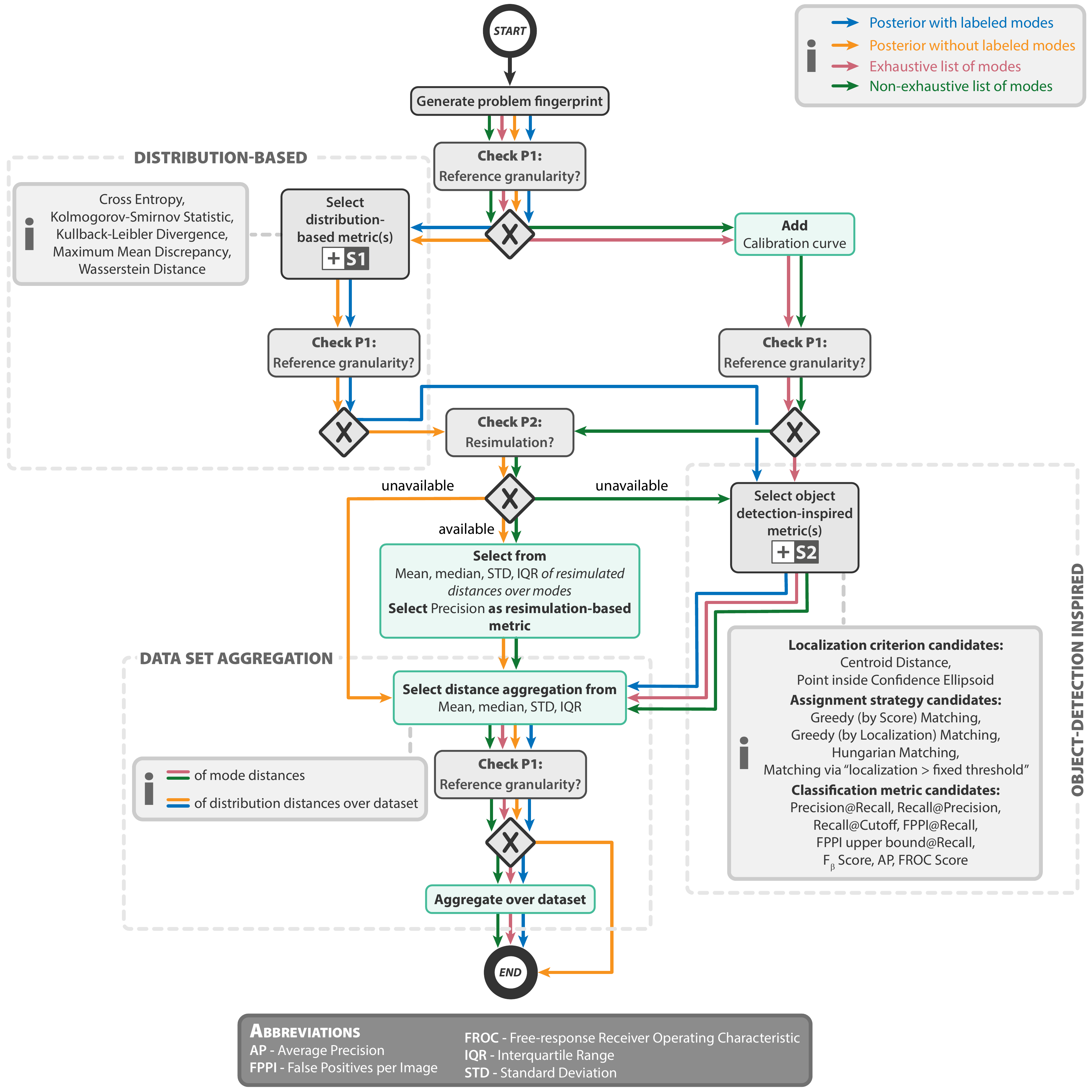}
    \caption{Overview of metric selection framework for posterior validation. Depending on the reference granularity (reference posterior with/without labeled modes, exhaustive or non-exhaustive list of reference modes), the user follows the correspondingly colored path in the decision tree. When a tree branches, the fingerprint items determine which exact path to take. Recommendations for distribution-based metrics (Subprocess S1) are provided in Fig.~\ref{fig:subprocess_1}. The main novelty of the proposal relates to the selection of object detection-inspired metrics, which is presented in a separate Subprocess S2 (Fig.~\ref{fig:subprocess_2}). The notation Metric1@Metric2 refers to providing the value for Metric1 for a specific target value (e.g. Recall = 0.95) of Metric 2. }
    \label{fig:meta_framework}
\end{figure*}

\section{Related Work}
\label{sec:related_work}

Prior work on recommendations for posterior validation is extremely sparse. While recent efforts have focused on recommendations in the context of classification, segmentation, and object detection \citep{maier2022metrics}, we have not found any framework dedicated to the validation of posteriors in inverse problems. Our analysis of the literature revealed the following common validation principles: (1) Use of the MAP as a point estimate and application of conventional regression metrics. This validation scheme is extended by regression metrics computed on resimulations (i.e., computing the forward model on the posteriors), if available. Furthermore, statistical distances to reference posteriors are commonly computed if the reference is actually given as a posterior. Lastly, visual inspection and qualitative analyses of the posterior (or interesting marginals) are also common practice (e.g., in \citep{ardizzone2018analyzing,haldemann2022exoplanet,butter2022two,kohl2018probabilistic,monteiro2020stochastic}). 

(2) In the context of conditional image generation, particular focus is put on the quality of the generated images and their diversity. This is reflected by commonly applied metrics such as peak signal-to-noise ratio (PSNR) or measures of variability (e.g., variance or standard deviation (STD)) of the generated images. At the same time, distribution-based metrics such as the Fréchet Inception Distance (FID) are also common but rarely applied to posteriors as a reference posterior is often lacking. Instead, validation or test images are interpreted as samples from an unconditional distribution and compared to samples drawn from the image generator. Depending on the exact image generation task, direct resimulation (e.g., in super-resolution tasks) or 'resimulation via a downstream task' (e.g., using an image classifier for class-conditioned image generators) might be an option, in which case such metrics are often reported (e.g., under the name of Consistency) \citep{ardizzone2019guided, ardizzone2021conditional, bao2023all, zheng2022entropy}.

To the best of our knowledge, a mode-centric validation has not been proposed before. Consequently, there is no prior work on using object detection validation methodology on posterior-based inverse problem solvers. 

\section{Methods}
\label{sec:methods}

This section presents our posterior validation framework (Sec.~\ref{subsec:problem_fingerprint}~-~\ref{sec. decision guide}) as well as the conditional Invertible Neural Network (cINN)-based architectures \citep{ardizzone2018analyzing,ardizzone2019guided} that we developed to instantiate the framework for medical vision problems (Sec.~\ref{sec:cinns}).

\begin{figure*}
    \centering
    \includegraphics[width=1\textwidth]{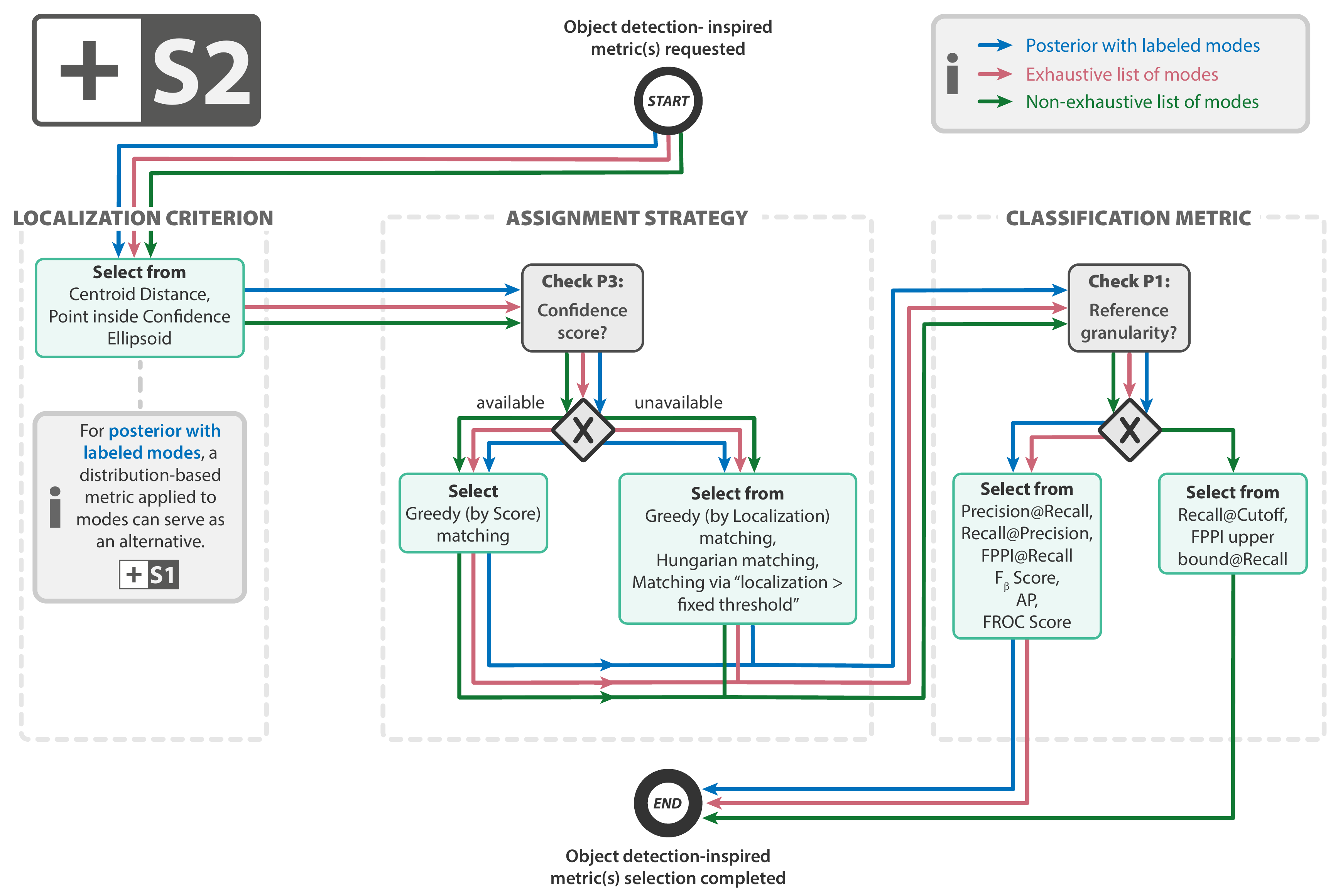}
    \caption{Subprocess S2 for selecting object detection-inspired metrics, comprising the steps of selecting the localization criterion, the assignment strategy, and the actual classification metric(s). The notation Metric1@Metric2 refers to providing the value for Metric1 for a specific target value (e.g. Recall = 0.95) of Metric 2. Decision guides for selecting a suitable option from a list of candidates are provided in section~\ref{sec. decision guide}. Used abbreviations: Average Precision (AP), Free-response Receiver Operating Characteristic (FROC), False Positives Per Image (FPPI).}
    \label{fig:subprocess_2}
\end{figure*}

Our validation framework features three main components to guide a user through the process of application-relevant metric selection. First, to enable an application-driven, modality-agnostic metric recommendation approach that generalizes over domains, we encapsulate validation-relevant characteristics of a given problem in a problem fingerprint. To this end, the parameters listed in Tab.~\ref{tab:fingerprint} are instantiated according to the domain interest. In a second step, suitable metrics are selected based on this problem fingerprint (Fig.~\ref{fig:meta_framework}). A key novelty in this step is the mode-centric validation perspective inspired by the field of object detection (Fig.~\ref{fig:subprocess_2}). Finally, as this process can result in a pool of suitable metric candidates, the third step involves the traversal of decision guides to help users understand the tradeoffs and choose between different candidates, wherever necessary. The following sections provide details on the three main components.

\subsection{Problem fingerprint}
\label{subsec:problem_fingerprint}
\begin{figure*}[htbp]
    \centering
    \includegraphics[width=1\textwidth]{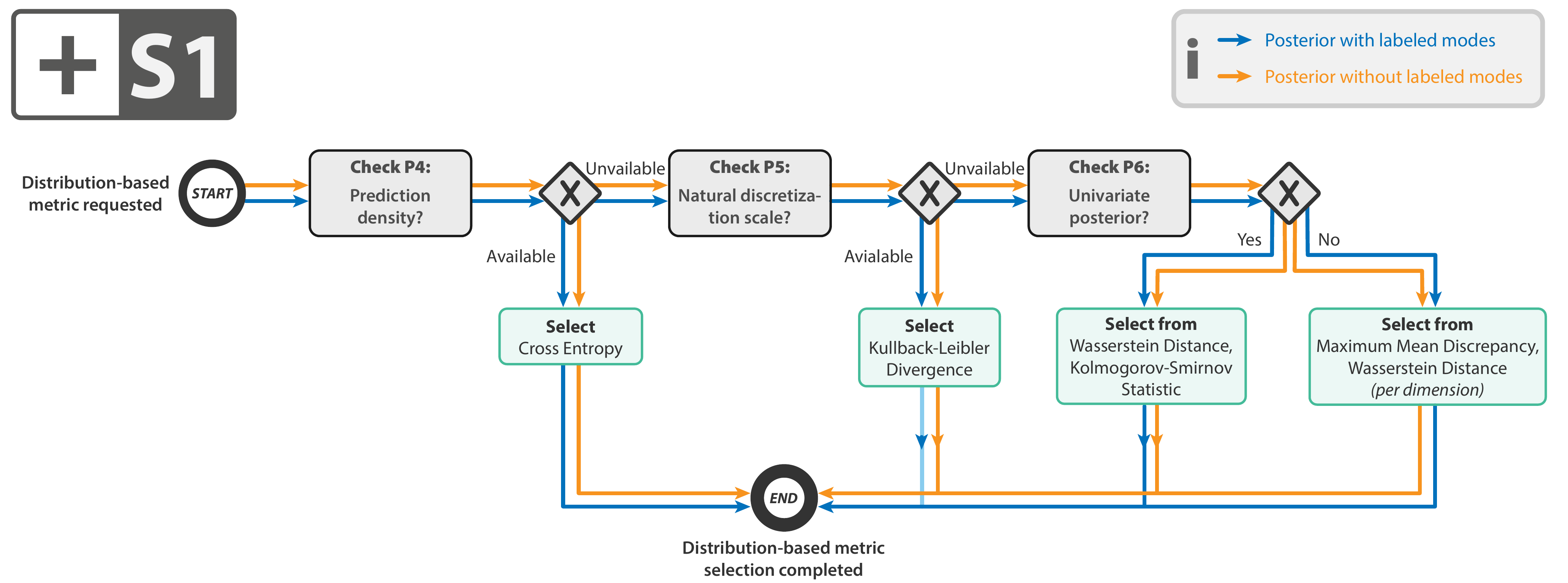}
    \caption{Subprocess S1 for selecting distribution-based metrics. Based on the exact representation of the predicted posterior and the dimensionality of the problem, different metrics become available.}
    \label{fig:subprocess_1}
\end{figure*}

The fingerprint is summarized in Tab.~\ref{tab:fingerprint}. While we assume the method to be validated to provide a posterior distribution, the framework can handle different types of references. Therefore, the most central fingerprint item is \textit{P1: Reference granularity} as it is the prerequisite for deciding whether distribution-based metrics and/or object-inspired metrics should be used for validation. Specifically, we distinguish four main formats in which the reference may be available (corresponding to the colored paths in Fig.~\ref{fig:meta_framework} and \ref{fig:subprocess_1}): posteriors with or without explicitly labeled modes, or a discrete set of modes that may either be exhaustive or non-exhaustive. Note that a non-exhaustive set of modes is very common in inverse problems because validation data is often generated with a forward model for which the underlying input serves as the (only) reference even if other inputs could have generated the same output (see Fig.~\ref{fig:use_case1_illustration}). Further properties will be detailed in the following.

\begin{longtblr}[
caption = {Full list of properties that comprise the inverse problem fingerprint. The first column contains the name of the property, the second column the values the property can take, and the third column a short description under which circumstances a value for a property should be chosen. Used abbreviations: conditional Invertible Neural Network (cINN), Generative Adversarial Network (GAN).},
label = {tab:fingerprint},
]{
    colspec={X[1,l,m]X[1, l, m]X[1.5, l, m]},
    }
    \hline[1pt]
    \textbf{Property} & \textbf{Possible Options} & \textbf{Description}\\
    \hline
    \SetCell[r=4]{t} {\textbf{P1} Reference granularity} & posterior with labeled modes & Choose this if the reference is a posterior including labels.\\
    & posterior without labeled modes & Choose this if the reference is a posterior but there is no mode information available.\\
    & exhaustive list of modes & Choose this if the reference is a discrete list of modes which is guaranteed to be complete.\\
    & non-exhaustive list of modes & Choose this if the reference is a discrete list of modes but modes might be missing on the list.\\
    \hline
    \SetCell[r=2]{t} {\textbf{P2} Resimulation} & available & Choose this if the forward model (often in the form of a simulation) is available and can be computed with acceptable resources.\\
    & unavailable & Choose this if the forward model is unavailable or prohibitively expensive.\\
    \hline
    \SetCell[r=2]{t} {\textbf{P3} Confidence score} & available & Choose this if your mode detection model can provide confidence scores for each mode individually. Ideally, the scores for modes of the same posterior are not innately correlated.\\
    & unavailable & Choose this if no suitable score is available.\\
    \hline
    \SetCell[r=2]{t} {\textbf{P4} Prediction density} & available & Choose this if your model provides an explicitly computable density function (e.g., cINNs).\\
    & unavailable & Choose this if your model represents the density implicitly (e.g., GANs).\\
    \hline
   \SetCell[r=2]{t} {\textbf{P5} Natural discretization scale} & available & Choose this if your application permits,
   e.g., a maximal necessary resolution.\\
   & unavailable & Choose this if your problem is, e.g., too high-dimensional (too many empty/singular bins) for discretization.\\
   \hline
   \SetCell[r=2]{t} {\textbf{P6} Univariate posterior} & yes & Choose this if the solution space of the inverse problem is 1D.\\
   & no & Choose this if the solution space of the inverse problem has two or more dimensions.\\
   \hline
   \SetCell[r=2]{t} {\textbf{P7} Accurate uncertainty required} & yes & Choose this if your application requires a well-calibrated uncertainty in the form of a correct posterior shape.\\
   & no & Choose this if accurate mode localization is sufficient.\\
   \hline[1pt]
\end{longtblr}

\subsection{Metric selection}
The workflow for metric selection, guided by the fingerprint, is provided in Fig.~\ref{fig:meta_framework}. The two main steps are:\\

\subsubsection{Selection of distribution-based metrics}
If reference posteriors are provided (Property P1), distribution-based metrics can be selected. The decision tree for selecting such a metric is depicted in Fig.~\ref{fig:subprocess_1}. The following properties are relevant in this context: 
    
\begin{itemize}
    \item \textit{P4: Prediction density:} Generative models can be categorized by whether they give access to the underlying density of the distribution they model (e.g., cINNs) or not (e.g., classical Generative Adversarial Networks (GANs) \citep{goodfellow2020generative}). There is also a grey area where the models provide bounds on the density (e.g., Variational Autoencoders (VAEs) \citep{kingma2013auto}). If the density is available, we can exploit it to gauge the mismatch between the predicted and reference distribution using the Cross Entropy \citep{good1956some}. The Cross Entropy needs access to the prediction density, but the reference density can be given as a sample. We propose the usage of Cross Entropy as it optimally exploits the availability of the density where it is accessible, whereas the other metrics make no explicit use of its existence.

    \item \textit{P5: Natural discretization scale:} Many problems allow for natural discretization, for instance, where there is a maximum necessary resolution for an application (e.g., 1 percentage point (pp) oxygenation resolution might be sufficient), and the range of the values is known. In such cases, the predicted and reference posteriors can be binned with acceptable discretization errors. Hence, the densities become mass functions, and the (discrete) Kullback-Leibler (KL) Divergence \citep{kullback1951information} is accessible. We propose this metric due to its lack of hyperparameters (except for the discretization parameters). However, if the solution space to the inverse problem is high 
    -dimensional, meaningful discretization is difficult due to the curse of dimensionality. In this case, we would encounter many empty bins and/or bins containing only a single sample. Such a binning is inadequate to estimate the probability mass function, and we discourage the use of the KL~Divergence.

    \item \textit{P6: Univariate posterior:} In some rare cases, we are interested in a single variable of interest as the solution to the inverse problem. If this is the case, the posterior will be univariate and there are statistical distances that tailor specifically to this setting. One example is the Kolmogorov-Smirnov (KS) statistic~\citep{kolmogorov1933sulla,smirnov1948table}, which gauges the difference between two univariate distributions based on their cumulative distribution function. The statistic itself can be used as a distance measure. Additionally, the KS statistic is the basis of a classical hypothesis test, which allows testing whether the posteriors significantly differ given some \(\alpha\)-level. An alternative to the KS statistic is the Wasserstein Distance~\citep{kantorovich1960mathematical}, which is defined for arbitrary dimensions but is computationally expensive for higher dimensions. Both distances have in common that they are almost free of hyperparameters (the KS test has the \(\alpha\)-level, and for the Wasserstein Distance, we have to choose the underlying L\(_p\)-norm with a tendency to choose \(p=1\) because the formula is particularly simple), which alleviates us of the necessity to 'optimize' the metrics on a validation data set. The Wasserstein Distance defines a metric (in the mathematical sense) on the space of distributions but does not directly lead to a hypothesis test in the same way the KS statistic does.
    
    \item \textit{P7: Accurate uncertainty required:} In contrast to the previous properties, which relate to "hard facts" about the inverse problem (such as the dimension of the solution space), this property is application-driven. In other words, whether we are interested in accurate uncertainty quantification does not depend on the underlying inverse problem but on the target application that requires solving the inverse problem. While not directly visible in the decision trees, the need for uncertainty quantification will influence the metric selection, for example, at the localization criterion (where we can decide to take a measure of variability of the modes into account). The influence is elaborated in the decision guides below. Additionally, the need for accurate uncertainty will inform the importance of the Calibration curve suggested as a metric for discrete reference modes in Fig.~\ref{fig:meta_framework}. 
\end{itemize}

If none of the properties P4 -- P6 lead to a suitable distribution-based metric, the user is left with two options (see Fig.~\ref{fig:subprocess_1}). The first is the Wasserstein Distance already introduced in the previous section. Its disadvantage is the computational cost in higher dimensions. A pragmatic solution is to apply the Wasserstein Distance to all 1D marginals individually and aggregate the results. This reduces the expressiveness of the Wasserstein Distance because there are distinct distributions with identical marginals, which could not be distinguished by this heuristic Wasserstein Distance. The other option is Maximum Mean Discrepancy (MMD)~\citep{gretton2012mmd}, which is a kernel method that introduces a metric on the space of distributions (at least for suitable kernels) and whose computational costs are acceptable. Its main downside is the sensitivity of the metric scores to the choice of the kernel (both the family and the hyperparameters parametrizing the family). This sensitivity often results in a separate validation set being required to optimize the hyperparameters of the metric and also reduces the interpretability of MMD.

Note that distribution-based metrics can also be used as a localization criterion when using object detection-inspired metrics which will be described in the following paragraph as depicted in Fig.~\ref{fig:subprocess_2}. \\

\subsubsection{Selection of object detection-inspired metrics}
If the reference comes with explicit modes, the quality of the modes should be explicitly assessed, possibly irrespective of the shape of the posterior (which is heavily influenced by the prior). We take inspiration from object detection validation by regarding predicted and reference modes as instances and transferring object detection principles to our setting. Our proposal is summarized in Fig.~\ref{fig:od_analogy}. 
\begin{itemize}
    \item Localization criterion: To decide whether a mode matches the reference, a criterion incorporating the location and (optionally) the shape of both reference and prediction is needed. Based on the application and goal, hyperparameters can be used to control the strictness of the criterion.
    \item Assignment strategy: To match the correct prediction/reference pairs, an adequate assignment strategy must be chosen. In this way, the matching of multiple predictions to one reference mode or vice versa is avoided.
    \item Classification metrics: Once predictions are located and assigned to reference predictions, a confusion matrix can be computed, and meaningful classification metrics can be calculated. Notably, these metrics should not rely on True Negatives (TNs), which are typically not well-defined in detection tasks. We argue that classification-based metrics offer a complementary view by treating modes - and thus the potential solutions to a problem - as the central objects of interest. 
\end{itemize}
Note that treating modes as instances introduces a hierarchy, where each posterior consists of one or more modes, and the data set consists of posteriors. This hierarchy should be respected during metric aggregation~\citep{maier2022metrics}. \\

To choose metrics for object-centric validation (if any), the following properties are of key importance:

\begin{itemize}
    \item \textit{P2: Resimulation (available/unavailable):} While the set of reference modes may be incomplete, it may be possible to verify whether a given mode (of the prediction) is another plausible solution to the problem. This can be achieved by applying the forward process (\textit{resimulation available}) to the given mode and choosing suitable metrics in the 'observable space'. The resimulation allows to decide whether a detected mode is a True Positive (TP) or False Positive (FP). With this information, the Precision, a highly relevant classification metric, can be computed.
    \item \textit{P3: Confidence score (available/unavailable):} Object detection metrics operating on the confusion matrix (e.g. the F\(_1\) Score) are highly sensitive to the method chosen to convert (fuzzy) algorithm output to actual decisions \citep{godau2023deployment}. Multi-threshold metrics, such as Average Precision (AP), overcome the need to decide on specific hyperparameters with ranking-based approaches. Transferring these principles to posterior validation requires the ability to rank the modes according to their likelihood of actually being a mode. This property should be set to true if the predicted modes come with a score that gauges the certainty of the model that the mode actually exists. While our framework is agnostic to the source of the score, we provide possible instantiations in our use cases in section~\ref{sec:cinns}. 
\end{itemize}

\subsection{Decision guides}
\label{sec. decision guide}
Our framework may result in users obtaining a pool of applicable metric candidates instead of only a single candidate. The decision guides presented in this section aim to help the user understand the tradeoffs between different metrics and selecting the most suitable candidate for their underlying problem. As many of the metrics are based on the observed object detection analogy, there are many parallels to the recommendations in \citep{maier2022metrics}.
The following paragraphs contain the decision guides for the ambiguous parts of the framework.

\begin{itemize}

    \item \textit{Localization criterion:} The localization criterion is used to gauge the agreement between pairs of predicted and reference modes. The choice of the localization criterion mainly depends on two properties: first, the granularity of the reference (P1, which is already covered in Subprocess S2 in Fig.~\ref{fig:subprocess_2}) and second, whether an accurate uncertainty is required (P7). If uncertainty quantification is important, the shape of the posterior modes should be taken into account when computing the mode localization. For a reference given as a mode location (without a spread or similar), this could take the form of computing the Mahalanobis Distance \citep{chandra1936generalised}, which takes the covariance of the predicted mode into account. This is an instance of the "Centroid Distance" category. The advantage of this metric is that it provides a continuous distance. On the other hand, the predicted mode could be used to construct a confidence ellipsoid (or a more general confidence region) to a given confidence level, and a match could be performed based on whether the reference location falls within the confidence ellipsoid ("Point inside Confidence Ellipsoid" category). This approach also takes uncertainty into account but leads to a binary score. If the reference is given as a distribution and accurate uncertainty is important, distribution-based metrics should be considered as these do not only match the mode location but incorporate the shape of the predicted and reference mode.

    If accurate uncertainty estimation is less important, the localization criterion should focus on the correct location of the mode centers. In this case, the predicted and reference mode should be collapsed to their centers, and a distance on these centers should be computed ("Centroid Distance" category). The exact distance should be chosen according to the application. Examples could be an L\(_p\)-norm for translation parameters, the cosine similarity \citep{schutze2008introduction} for rotational variables, or structural similarity index \citep{wang2004image} for images.

    \item \textit{Assignment strategy:} Whenever an uncertainty score is available, greedy matching via the (confidence) score \citep{maier2022metrics,everingham2015pascal} should be applied. The rationale behind this recommendation is that models that confidently predict wrong or far-off modes should be penalized. If no confidence score is available, there are multiple complementary options. Greedy matching via the localization criterion \citep{maier2022metrics} has the advantage of being methodologically simple and computationally fast. Furthermore, depending on the application, it can be sensible to match the closest modes first. An alternative would be to apply Hungarian matching \citep{maier2022metrics,kuhn1955hungarian}, which finds an optimal matching that minimizes the total mode distances. Such a matching can lead to a predicted mode not being matched with its closest reference mode. Hungarian matching can be suitable for a more theory-focused validation or method comparison (independent of a downstream application). However, as elaborated in \citep{maier2022metrics}, Hungarian matching can lead to overly optimistic assignments, artificially reducing the number of FNs and FPs. Lastly, assigning modes via a fixed localization threshold ("Matching via Localization \(>\) Fixed Threshold") can be useful if the application requires an exact number of predicted modes but less focus on the precise localization of the modes. An example downstream task would be to count the occurrence of certain structures.

    \item \textit{Distance aggregation:} An important aspect of distance aggregation is to respect the hierarchical structure of the data, as elaborated in \citep{maier2022metrics}. In this posterior-based inverse problem setting, a data set consists of data points, where each data point corresponds to a set of reference modes and a set of predicted modes. This two-stage hierarchy implies that first, the distances between modes per posterior should be aggregated before these per-data-point distances should be aggregated over the whole data set. In Fig.~\ref{fig:meta_framework}, we explicitly mention mean, median, STD, and Interquartile Range (IQR) as aggregation methods for distance aggregation. However, these solely represent examples of common choices. Depending on the application, it might be advantageous to report other quantiles of the distribution (instead of IQR) or weight the data points in the mean. Overall, it should be noted that quantile-based aggregates (such as median or IQR) are more robust to noise and outliers, which might make them superior to mean and STD, as many models produce rather noisy posteriors.

    \item\textit{Classification metrics:} If a confidence score is available, we recommend multi-threshold metrics such as AP or FPPI in almost all cases. They address the problem of noisy modes due to imperfections in the posterior generation and/or clustering methods. Metric@(TargetMetric = TargetValue), as introduced in \citep{maier2022metrics}, is a notation to report the value of a metric while a target metric is optimized on a dedicated validation split to conform to the target value. An example would be Precision@(Recall=0.95). This type of metric should be chosen if the application requires certain bounds, e.g., on the frequency of FPs, as might for instance be derived from regulatory requirements. Reporting of this form is also common practice in clinically-focused communities. F\(_\beta\) \citep{maier2022metrics,van1979information,chinchor1993muc} aggregates both Precision and Recall and can be useful if there is no target value for either one, but instead, the model (hyper-)parameters are optimized (on an additional validation set) to maximize F\(_\beta\). 
\end{itemize}

\subsection{Conditional Invertible Neural Networks for ambiguous problems}
\label{sec:cinns}

To showcase the benefit of our framework, we investigate three complementary inverse problems that feature inherent ambiguity (see Figs.~\ref{fig:toy_example} -- \ref{fig:use_case_2}). These problems exhibit either mathematical ambiguity, where multiple solutions are exactly correct due to the problem's structure, or practical ambiguity, where solutions become effectively indistinguishable due to measurement noise and system limitations. In the following, we present these use cases along with the methods whose performance is to be assessed with our framework. Further implementation details can be found in  ~\ref{sec:implementation_details}. 

\subsubsection{Toy example}
\begin{figure*}
    \centering
    \includegraphics[width=1\textwidth]{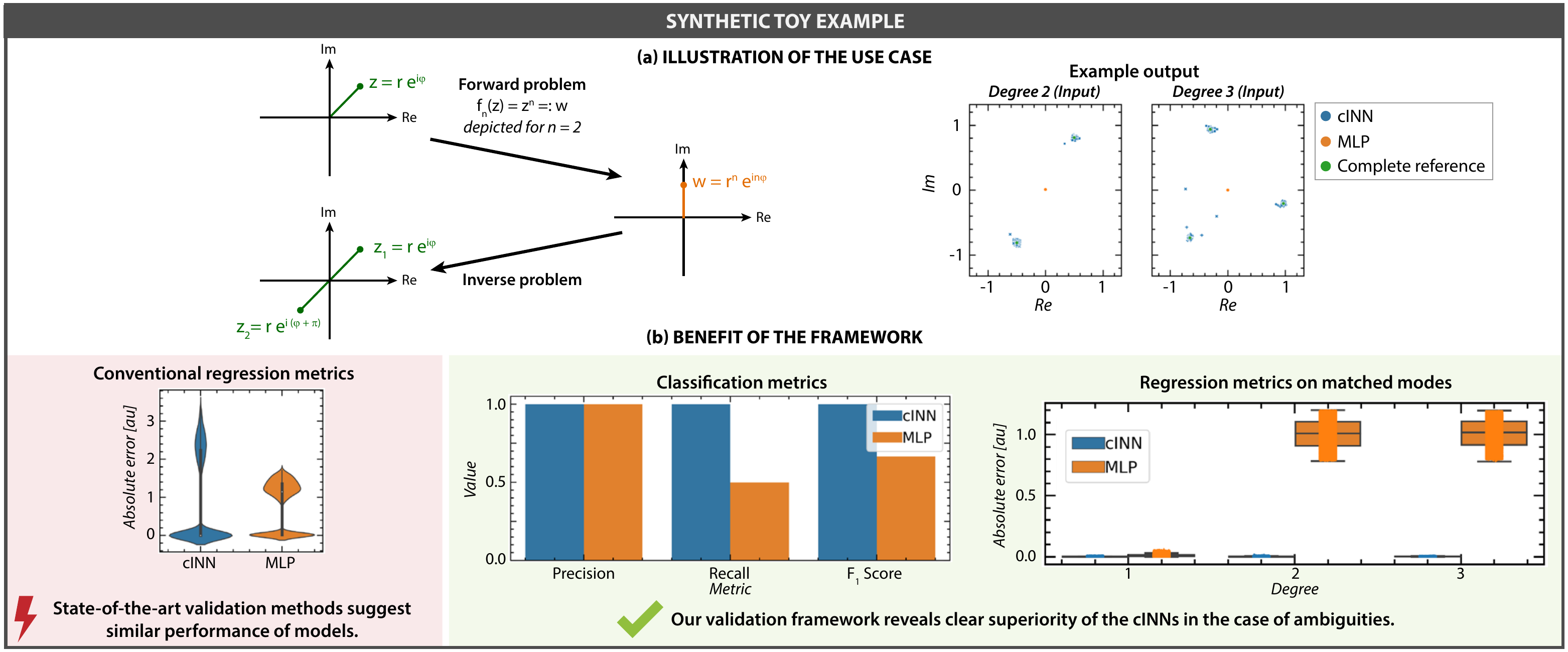}
    \caption{Results for the synthetic toy example. (a) The task consists of computing the \(n\)th root(s) (\(n=1, 2, 3\)) of a non-zero complex number. While the conditional Invertible Neural Network (cINN) captures the ambiguity of the problem via multiple modes in the posterior, a classical multi-layer perceptron (MLP) typically outputs the mean of plausible solutions. (b) Left: The superiority of the cINN is not captured by conventional validation methods that treat the problem as a regression task (assuming a unique solution) using the maximum a posteriori probability as the cINN estimate. Right: The explicit mode localization and assignment offered by our framework enables the computation of classification metrics and regression metrics applied on matched modes. These reveal the poor performance of the MLP compared to the cINN.}
    \label{fig:toy_example}
\end{figure*}

As a toy example, we chose a well-understood, but ambiguous, inverse problem, namely finding the \(n\)-th roots of a complex number \(w\) for varying \(n\) (cf. Fig.~\ref{fig:toy_example} (a), left). The input to the inverse problem is the complex number \(w\) for which to find the root(s) and the integer \(n\) describing the order of the root.
We considered two models: (1) A multi-layer perceptron (MLP) (based on \citep{kendall2017uncertainties}) as a naive baseline, which, given \(n\) and \(w\), produces a Gaussian posterior represented by a mean and a diagonal covariance matrix. (2) A cINN \citep{ardizzone2019guided}, which, given \(n\) and \(w\), produces a posterior distribution over \(z\) by sampling a latent space. As a mode detection algorithm, we used the clustering algorithm Density-Based Spatial Clustering of Applications with Noise (DBSCAN) \citep{ester1996density}. To estimate the MAP probability location, we used the mean of the largest cluster. 

\begin{figure*}
    \centering
    \includegraphics[width=1\textwidth]{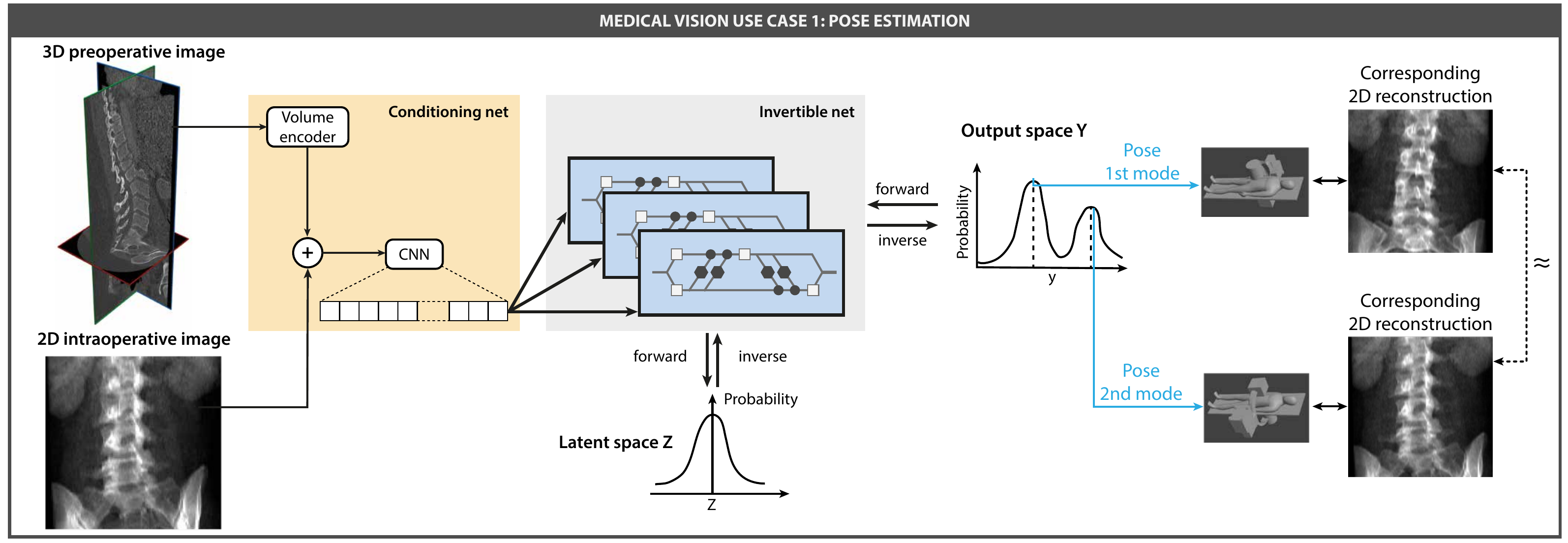}
    \caption{Example of an inverse problem in medical vision. The ambiguity of the problem can be captured with an invertible architecture, which represents multiple solutions (here: two) via modes in a posterior distribution. Used abbreviation: Convolutional Neural Network (CNN).}
    \label{fig:use_case1_illustration}
\end{figure*}

\subsubsection{cINNs for pose estimation in intraoperative 2D/3D registration}
\label{sec:methods_registration}
Image registration is the basis for many applications in the fields of medical image computing and computer-assisted interventions. One example is the registration of 2D X-ray images with preoperative 3D CT images in intraoperative surgical guidance systems, as illustrated in Fig.~\ref{fig:use_case1_illustration}. Previously proposed methods~\citep{registration_dl,salehi2018real,miao2016,yang2016} lack the capacity to represent the inherent ambiguity a registration problem may contain, i.e., they cannot handle a situation where multiple substantially different solutions exist. Based on our own preliminary work, we address this lack with cINNs, by representing the possible solutions to a registration problem through a non-parametric probability distribution that encodes different plausible solutions via multiple modes \citep{trofimova2020representing}. The challenge of detecting modes in high-dimensional parameter space is tackled by interpreting the task as a clustering problem performed on the samples defining the posterior. The neural network architecture is illustrated in Fig.~\ref{fig:use_case1_illustration}. The input images are passed through a conditioning network such that a relatively low-dimensional vector (here: 256) can be used for conditioning the actual invertible net. 

\subsubsection{cINNs for quantification of functional tissue parameters}
\label{sec:methods_pat}
Photoacoustic imaging is an emerging modality that enables the recovery of functional tissue parameters. However, the underlying inverse problems are ill-posed (Fig.~\ref{fig:use_case_2}). Specifically, the problem might have ambiguous solutions, meaning that different tissue compositions could lead to the same photoacoustic measurement. We address this ambiguity with a cINN-based architecture as proposed in \cite{nolke2021invertible, nolkePhotoacousticQuantificationTissue2024}. As a naive baseline, we chose the state-of-the-art method "Learned Spectral Decoloring" (LSD) \citep{grohl2021learned} based on a fully connected neural network architecture, which provides a single point estimate as a prediction. We optimized the architecture and training procedure for better performance. For clustering, we used the UniDip Clustering algorithm \citep{maurus2016skinny}, which is based on the Hartigan-Dip test for unimodality \citep{hartigan_dip_1985}. It provides robust estimations with respect to resampling of the posterior and is basically parameter-free (apart from a statistical significance level). 

\section{Experiments \& Results}
\label{sec:experiments}
The purpose of the experiments was to instantiate our framework for several use cases and showcase the added value by means of examples. All validations were performed on independent test sets. However, note that we did not aim to optimize the models for the use cases or solve the underlying tasks. Instead, the focus was on the insights that can be derived from the proposed validation scheme.

\begin{figure}[htbp]
    \centering
    \includegraphics[width=.6\textwidth]{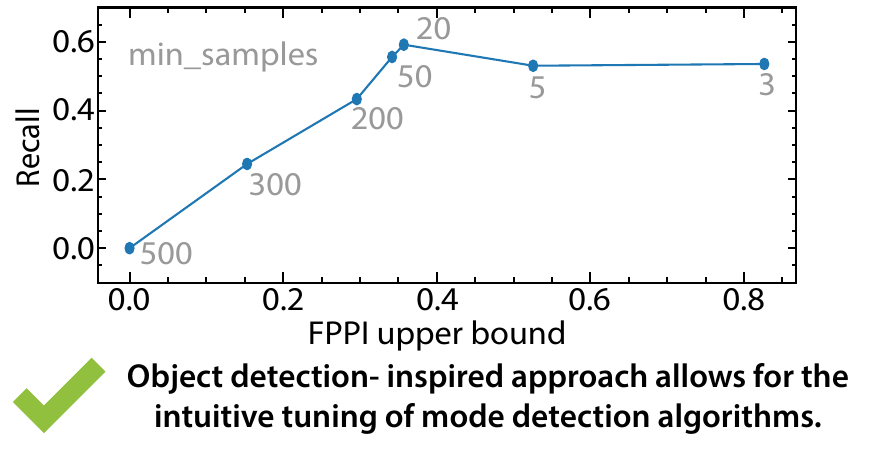}
    \caption{Use case: pose estimation in surgery. Mode detection algorithms can be configured to provide either higher Recall (at the cost of more False Positives (FPs)) or higher Precision (at the cost of more False Negatives (FNs)). Our framework captures this tradeoff by performing explicit mode localization/matching and recommending the plotting of the Recall as a function of the FPs per image (FPPI).}
    \label{fig:use_case_1_results}
\end{figure}

\subsection{Synthetic toy example}
\label{sec:experiments_toy}

The purpose of the toy experiment was to validate the framework branch for a reference with an exhaustive list of modes. As described in Sec.~\ref{sec:methods}, the task was to, given an integer \(n\) and a complex number \(w = R\cdot e^{i\Phi}\), compute the $n$th root of \(w\). The distinct solutions (assuming \(w \neq 0\)) to this inverse problem can be explicitly enumerated as \(z_k = \sqrt[n]{R} \cdot e^{i\frac{\Phi+ 2\pi k}{n}}\) for \(k=0, \dots, n-1\). The training data consisted of tuples \((z, n, w)\), such that \(z^n = w\). To highlight the pitfalls of treating such a problem as a simple regression task (assuming a unique solution), we trained an MLP and a cINN to estimate \(z\) from \(w\) and \(n\) and evaluated their performance using the absolute error. Specifically, for each test sample, the absolute error was calculated between the model's prediction and the original \(z\) that was used to generate that particular \((z, n, w)\) tuple. Additionally, we instantiated our framework, which provided us with additional metrics taking the number of modes and the matching process into account. These additional metrics were Precision, Recall, \(F_\beta\) (we report \(\beta = 1\)), AP, and the absolute error computed for matched modes and aggregated per posterior. The ranking of modes, required for the computation of AP was achieved by bootstrapping the posteriors. More specifically, we resampled each posterior two times and computed the Intersection over Union (IoU) of the new clusters with the original clustering. The average IoU per cluster was used as confidence score.

To instantiate our training and testing sets, we drew \(n\) uniformly from the set \(\{1, 2, 3\}\), \(z\) uniformly from an annulus centered at \(0\) with inner radius \(0.8\) and outer radius \(1.2\), and `simulated' the forward process via \(w = z^n\). The training set consisted of \(10^6\) samples, and the testing set of \(10^5\) samples. \(n\) was one-hot encoded, and \(z\) and \(w\) were represented using their real and imaginary part, respectively. We used 1024 latent samples to build the posterior during inference. As a localization criterion, we chose the mode center distance. The predicted and reference modes were matched greedily by the assignment strategy.

Fig.~\ref{fig:toy_example} (b) left shows the conventional absolute error distribution of the two models, indicating that while both models perform poorly, the MLP might be superior to the cINN. This is in contrast to qualitative observations (example shown in  Fig.~\ref{fig:toy_example} (a)), which suggest good performance of the cINN, while the MLP seems to predict the mean of the ambiguous solutions (which is \(0\)). 

The framework metrics unmask this performance difference between the models. While both models perform similarly regarding Precision, the cINN outperforms the MLP in terms of Recall, which is due to the fact that the cINN is capable of predicting multiple modes, while the MLP is restricted to a single mode. The absolute error of the matched modes underlines that for higher-order roots, the cINN correctly identifies the locations of the root, while the MLP predictions are only close to the ground truth in the unambiguous case of \(n=1\). The cINN achieved an AP of approximately 1.

\subsection{Medical vision use case 1: Pose estimation}
\label{sec:experiments_pose}

To showcase the potential of the framework for model optimization, we picked a surgical use case (Fig.~\ref{fig:use_case1_illustration}). In this setting, ambiguity in pose estimation results from the general symmetry of the spine and limited image quality. To generate a validation data set with reliable references, we simulated X-ray images taken by a C-Arm with multiple orientations using the principle of digitally reconstructed radiographs (DRRs) \citep{unberath2018deepdrr}. As our experimental data set, we used the UWSpine data set~\citep{dataset1,dataset2}, which comprises spine-focused CT volumes of 125 patients. We transformed the volumes to a homogeneous voxel spacing and discarded images smaller
than 128x256x128 as well as patients with an asymmetric spine. For every CT volume, we sampled a set of different poses of the C-Arm device and computed corresponding DRRs. We split the data into a disjoint training and test data set (no overlap between patients) with 131,900 and 2,700 samples, respectively. For our validation, we only considered samples with a highly symmetric spine, which resulted in 196 samples. For each posterior we sampled from the latent space 1,028 times.

From an application perspective, the conversion of posteriors to modes (i.e., the actual solutions of interest) is a crucial step in the system. Often, mode detection algorithms can be configured to provide either higher Recall (at the cost of more FPs) or higher Precision (at the cost of more False Negatives (FNs)). To address this tradeoff, we applied our framework for hyperparameter tuning. Based on the suggested mode matching, we plotted the Recall (using only the reference modes provided by the simulation) as a function of the (upper bound of the) FPPI for different hyperparameters of the mode clustering algorithm. Note that we speak of an upper bound because of the non-exhaustive list of modes. We varied the minimum samples parameter of the DBSCAN algorithm. Given that we only worked with symmetric spines, we regarded a mode corresponding to a left anterior oblique (LAO) angle of LAO\(_\text{ref} + 180\)° as a TP. Based on the recommendation framework (Figs.~\ref{fig:meta_framework} / \ref{fig:subprocess_1}), we chose the Centroid Distance as the localization criterion (threshold 20°) and Greedy by Localization as the assignment strategy. Fig.~\ref{fig:use_case_1_results} reveals that the cluster algorithm hyperparameters corresponding to an FPPI of approximately 0.35 provide the best tradeoff. This analysis was enabled by the detection-driven validation approach.

\subsection{Medical vision use case 2: Functional tissue parameter estimation}
\label{sec:experiments_pat}

\begin{figure*}
    \centering
    \includegraphics[width=1\textwidth]{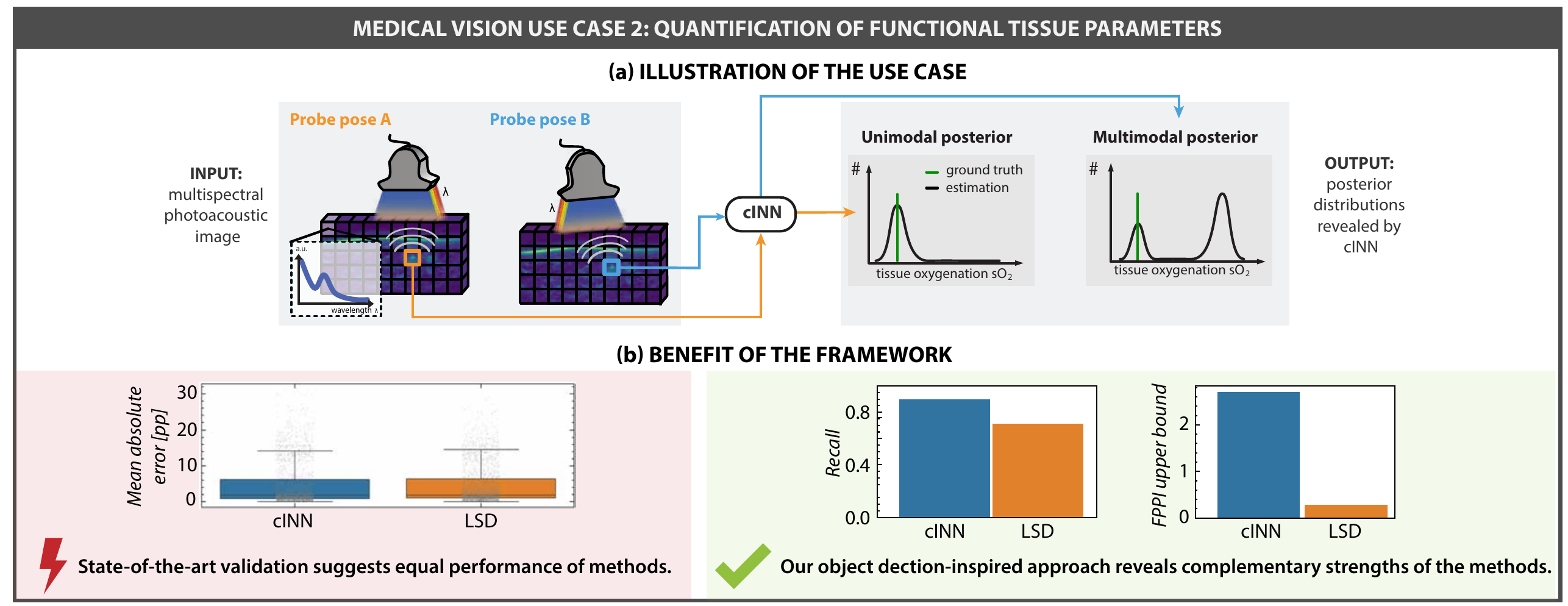}
    \caption{Use case: functional tissue parameter estimation. (a) The task is to estimate blood oxygenation (sO\(_2\)) from multispectral photoacoustic imaging data. The potential ambiguity of the problem for a given location (e.g., a vessel) can be resolved by changing the pose of the image modality (pose 1: unique solution; pose 2: multiple plausible solutions). (b) Left: Conventional validation methods based on maximum a posteriori estimates do not capture the full capabilities of different methods, specifically not the advantageous properties of conditional Invertible Neural Networks (cINNs) compared to conventional point estimation methods such as "Learned Spectral Decoloring" (LSD). Right: The explicit mode localization and assignment offered by our framework enable the computation of classification metrics. These reveal the application-relevant properties of the methods, namely the Recall and the False Positives Per Image (FPPI).}
    \label{fig:use_case_2}
\end{figure*}

The second medical vision use case is illustrated in Fig.~\ref{fig:use_case_2} and concerns the quantification of tissue oxygenation from photoacoustic measurements. The purpose of this experiment was to demonstrate that common validation methods are not well-suited for application-driven performance assessment. To this end, we trained two models (see Sec.~\ref{sec:methods}) for the given use case -- a naive baseline that treats the problem as a regression problem with a unique solution as well as a solution based on cINNs. Since ground truth tissue properties are unavailable for in vivo photoacoustic measurements, we simulated a synthetic data set of human forearm images using the Monte Carlo method for light propagation (Monte Carlo eXtreme (MCX)). Our digital tissue model was inspired by the work of \cite{schellenberg2022photoacoustic} and consists of different tissue types (skin, muscle background, arteries, veins, ultrasound gel, membrane, and water). The anatomic and optical properties are based on knowledge from literature. The whole simulation was implemented using the \textit{Simulation and Image Processing for Photonics and Acoustics} (SIMPA) toolkit \citep{grohl2022simpa}.A total of 1100 synthetic photoacoustic images of the human forearm were simulated (Train:Val:Test; 900:100:100 images) \citep{schellenberg2022photoacoustic}. For each posterior 5,000 samples were drawn during inference time.
For our validation, we focused on samples that were detected to be multimodal by our cINN. As a first naive validation approach, we compared the MAP estimate, i.e., for the cINN, we used the median of the largest cluster as a point estimate. As can be seen in Fig.~\ref{fig:use_case_2} (b) left, both methods seem to perform equally well. Note that for bimodal posterior distributions, the MAP is not necessarily the best solution, as sometimes the smaller mode might correspond to the reference (in this manuscript, we refer to ``large'' and ``small'' modes based on the number of samples that were grouped in one cluster). Point prediction methods such as LSD usually predict a value that is either close to the largest mode of the cINN or lies between the two modes (as in the toy example). Our framework addresses this issue. Following the recommendation framework, we first performed mode matching (Greedy by Localization) with a threshold of 5 percentage points (pp) sO\(_2\) difference to enable object detection-inspired metrics. In analogy to the previous example, we then computed Recall and FPPI upper bound. The cINN method outperforms LSD in terms of Recall (90\% vs. 71\%). However, this comes at the cost of more FPPI. 

\section{Discussion and conclusion}
\label{sec:discussion}
Validation of deep learning-based methods attempting to solve inverse problems is key for both measuring progress as well as their eventual translation to real-world applications. Currently, however, common validation practice frequently neglects the requirements of the underlying application, leading to the resulting metric scores often not reflecting the actual needs. This especially holds true for posterior-based methods tackling inverse problems for which multiple different but plausible solutions exist. 

Currently, inverse problem solvers, whether using a posterior or classical representation, are often validated in an ad hoc manner specifically tailored to the problem at hand. Our posterior validation framework takes one step back and proposes key properties that allow us to abstract from the specific inverse problem and advance toward a unified, generic inverse problem validation methodology. As we argue that flaws in common validation practice can largely be attributed to a lack of best practices, in our opinion, dedicating efforts towards improving common practice becomes imperative to advance the field. Our framework provides a first step towards structured and standardized validation practice. We hope that an according shift in research focus exemplified by our work sparks further research on how to best validate inverse problem methods and allows for better and more meaningful comparisons of algorithms.

With the proposed framework, we are -- to the best of our knowledge -- the first to systematically address this problem. A particular novelty is the leveraging of object detection-inspired metrics for posterior validation, which enables a mode-centric validation. The mode-centric view aligns naturally with applications, for example in the medical domain, where interpretation of a posterior distribution might be infeasible, but the scanning of a (short) list of plausible solutions might provide a benefit over a point prediction both in terms of predictive performance as well as uncertainty quantification.

While a direct evaluation of our proposed framework is not possible, we instead demonstrated its value in various medical vision use cases that have multiple disjunct plausible solutions. Further use cases include inverse kinematics \citep{ardizzone2018analyzing,amesIKFlowGeneratingDiverse2022} and robotic grasping \citep{ghazaeiDealingAmbiguityRobotic2019} in robotic surgery, tumor volume assessment in CT imaging \citep{kohl2018probabilistic, rahmanAmbiguousMedicalImage2023}, and object pose estimation \citep{manhardtExplainingAmbiguityObject2019a, jamesjensenNovelPostprocessingTechnique2024}. However, some of the emerging applications of posterior-based deep learning methods, such as general image reconstruction problems, may exhibit modes that are less disjunct than in these examples, and it remains to be seen whether the object detection-inspired metrics will be amenable to such scenarios. Note that for the presented examples neither the models used nor the experimental data have been optimized for the particular use case as our focus was on demonstrating the framework's capabilities rather than solving specific clinical problems.

A limitation of our mode-centric validation strategy is that it relies on clustering algorithms, which face challenges in higher dimensions \citep{steinbach2004challenges}. However, the manifold hypothesis suggests that high-dimensional data often lies close to a lower-dimensional manifold \citep{feffermanTestingManifoldHypothesis2016} and recent work has demonstrated the effectiveness of learned latent representations through autoencoders in handling such data \citep{rombachHighResolutionImageSynthesis2022}.
Multi-threshold metrics such as AP are widely used metrics in object detection and, as such, are also included in our framework. However, it must be noted that a critical requirement for their computation is the availability of a confidence score. Natural choices for confidence scores such as the relative mass of the mode have disadvantages such as the confidence score depending on the number of detected modes. Future work should thus be directed toward developing alternative confidence scores overcoming this limitation and enabling the use of these robust metrics. Also, a future implementation of the metrics in a library will be useful in providing the community with a standardized and reliable resource for validation, given that previous work highlighted the problems of non-standardized metric implementation \citep{maier2022metrics, reinke2023understanding}.
On a further note, the pool of available metrics in the case of non-exhaustive reference modes is currently rather limited. This is because only two of the four binary confusion matrix entries are available (TP and FN). This limitation particularly affects applications where we cannot evaluate a forward model or where the evaluation is prohibitively expensive. One example is data sets consisting of observational data. In this case, there is often no way to perform an intervention to test whether a predicted mode is a plausible solution. Other examples are Monte Carlo-based forward models, as these take a long time to compute a stable solution. While it may be possible to check a small number of predictions using these forward models, systematically checking each prediction in a large test set will very quickly become computationally expensive. We hope that the clear structure using the inverse problem fingerprints will spark a fruitful discussion on new metric candidates suitable for this setting. 

In conclusion, our experiments clearly demonstrate the added value of mode-centric validation compared to the standard validation approach. Our framework could thus evolve as an important tool for posterior validation in inverse problems.

\section*{CRediT authorship contribution statement}
\textbf{Tim J. Adler:} Conceptualization, Methodology, Software, Validation, Formal analysis, Investigation, Data Curation, Writing - Original Draft, Writing - Review \& Editing, Visualization. \textbf{Jan-Hinrich Nölke:} Conceptualization, Methodology, Software, Validation, Formal analysis, Investigation, Data Curation, Writing - Original Draft, Writing - Review \& Editing, Visualization.\textbf{Annika Reinke:} Conceptualization, Methodology, Validation, Writing - Review \& Editing, Visualization. \textbf{Minu Dietlinde Tizabi:} Validation, Writing - Original Draft, Writing - Review \& Editing. \textbf{Sebastian Gruber:} Software, Validation, Formal analysis, Investigation, Data Curation, Writing - Review \& Editing. \textbf{Dasha Trofimova:} Software, Validation, Formal analysis, Investigation, Data Curation, Writing - Review \& Editing. \textbf{Lynton Ardizzone:} Software, Writing - Review \& Editing. \textbf{Paul F. Jaeger:} Conceptualization, Methodology, Writing - Review \& Editing, Funding acquisition. \textbf{Florian Buettner:} Conceptualization, Methodology, Writing - Review \& Editing, Funding acquisition. \textbf{Ullrich Köthe:} Conceptualization, Methodology, Writing - Review \& Editing. \textbf{Lena Maier-Hein:} Conceptualization, Methodology, Resources, Writing - Original Draft, Writing - Review \& Editing, Supervision, Project Administration, Funding acquisition.
\section*{Declaration of competing interest}
The authors declare that they have no known competing financial interests or personal relationships that could have appeared to influence the work reported in this paper.
\section*{Data availability}
Data used in this paper are publicly available or can be accessed upon request. We have cited all the corresponding sources.
\section*{Acknowledgments}
The authors would like to thank Alexander Baumann, Kris Dreher, Alexandra Ertl, Julius Holzschuh, Marc Kachelrieß and Alexander Seitel for valuable discussions, Marcel Knopp for proofreading as well as Melanie Schellenberg and Tom Rix for their contributions to figure design. This project has received funding from the European Research Council (ERC) under the European Union’s Horizon 2020 research and innovation programme (NEURAL SPICING: grant agreement No. [101002198]). Part of this work was funded by Helmholtz Imaging (HI), a platform of the Helmholtz Incubator on Information and Data Science. The work was co-funded by the European Union (ERC, TAIPO, 101088594). Views and opinions expressed are however those of the authors only and do not necessarily reflect those of the European Union or the European Research Council. Neither the European Union nor the granting authority can be held responsible for them.
\section*{Declaration of generative AI and AI-assisted technologies in the writing process}
During the preparation of this work, the authors used Claude 3.5 Sonnet through Perplexity AI and DeepL Write in order to assist with language editing and text refinement. After using this tool/service, the authors reviewed and edited the content as needed and take full responsibility for the content of the published article.

\appendix
\section{Use case implementation details}
\label{sec:implementation_details}

The following sections describe the implementation details of the models used to solve the inverse problems in the use cases. All deep learning models were implemented using PyTorch. The cINNs further made use of the Framework for Easily Invertible Architectures (FrEIA) \citep{freia}. \\

\subsection{Synthetic toy example}
\begin{itemize}

    \item \textit{Architecture:} The MLP was implemented following the architecture proposed in \citep{kendall2017uncertainties}. More precisely, we implemented a six-layer fully-connected neural network with rectified linear unit (ReLU) activations, 128 dimensions for each hidden layer, and a dropout rate of 0.2. The output was four-dimensional, and we interpreted the first two dimensions as the mean and the second two dimensions as the logarithmic standard deviation of a Gaussian distribution with diagonal covariance over the solution space (i.e., the space of possible roots). The network was trained using maximum likelihood training under the Gaussian assumption, which corresponds to the loss

\[
L = \mathbb{E}_{(z, n, w)}\left [ \frac12 \sum_{i=1}^2 \left( e^{-2\delta_i} \cdot ( z_i - \hat z_i)^2 + 2\delta_i + \log(2\pi) \right) \right],
\]
where \(\hat z, \delta = f_\Theta(w, n)\) are the model predictions. We applied Monte Carlo dropout at inference time, which led to multiple predictions \((\hat z(k), \delta(k))_{k=1,\dots,N}\), which we aggregated via

\begin{align*}
\hat z & = \frac1N \sum_{k=1}^N \hat z(k),\\
\hat \sigma_i(k) & = e^{\delta_i(k)}, \text{ and}\\
\hat \sigma_i & = \sqrt{\frac1N \sum_{k=1}^N \hat z_i(k)^2 - \hat z_i^2 + \frac1N \sum_{k=1}^N \hat \sigma_i(k)^2},
\end{align*}
for \(i \in \{1,2\}\) denoting the axes of the standard identification of \(\mathbb{R}^2\) with the complex numbers via real and imaginary part.

The model was trained for 1000 epochs using the AdamW \citep{loshchilov2017decoupled} optimizer, with a learning rate of \(10^{-3}\), a weight decay parameter of \(10^{-5}\), and a batch size of 2048. For inference, we chose \(N=50\) following \citep{kendall2017uncertainties}.

The cINN was implemented using affine coupling blocks \citep{dinh2014nice,dinh2016density,ardizzone2018analyzing,kingma2018glow} followed by (fixed) random permutations and a global affine transformation (i.e., an affine transformation with learnable parameters but independent of the input to it) \citep{kingma2018glow,mackowiak2021generative}. We used 20 affine coupling blocks with shallow fully-connected subnetworks with a single hidden layer with 256 dimensions and ReLU activations. The scaling of the affine coupling was soft-clamped with a clamping constant of 2.0, and we initialized the global affine transformation with the scaling parameter 0.7 (in FrEIA this parameter is called global\_affine\_init). The cINN works by transforming the solution of the inverse problem (z in our case) into a latent space, conditioned on the observables (w and n), i.e., the cINN is a map g(z; w, n) that is invertible with regard to z, given w and n. During training, a Gaussian distribution on the latent space is enforced via maximum likelihood training:

\[
L = \mathbb{E}_{(z, n, w)} \left [ \frac12 \sum_{i=1}^2 g_i(z; n, w)^2 - \log |\det Jg(z; n, w)| \right],
\]
where \(Jg\) denotes the Jacobi matrix of \(g\). The architecture of the cINN is chosen in such a way that the Jacobi matrix is triangular, such that the log-determinant is efficiently computable. At inference time, we draw samples in the latent space and transform them to the solution space via \(g^{-1}\) (given w and n). 

The cINN was trained for 1000 epochs using the AdamW optimizer with a learning rate of \(10^{-2}\), which was reduced by a factor of 10 after epochs 200, 500, and 900. We used a weight decay parameter of \(10^{-5}\) and a batch size of 2048. 

Before training, z and w were normalized to zero mean and unit variance. The one-hot encoded n was left unchanged. Furthermore, we applied noise augmentation with a standard deviation of 0.02 to the normalized z and w dimensions.

    \item \textit{Mode Processing:} For mode detection of the cINN posteriors, we applied the DBSCAN \citep{ester1996density} clustering algorithm using the scikit-learn library. DBSCAN was applied to the denormalized data with a minimum sample size of 20 and \(\varepsilon = 0.2\). 
\end{itemize}

\subsection{Medical vision use case 1: Pose estimation}

\begin{itemize}

    \item \textit{Data Set:} For every CT volume, we sampled 100 different poses of the C-Arm device and computed corresponding DRRs. The virtual C-Arm poses (relative to the 3D volume coordinate system) were determined as follows: The translation along the sagittal, longitudinal, and transverse axis was randomly sampled from a continuous uniform distribution with range [-20 mm, 20 mm]. The two angles representing the rotation around the longitudinal (LAO) and transverse (CRAN) axis of the patient were sampled from a discrete uniform distribution with range [-\ang{20}, \ang{20}] and a step size of \ang{1}. With a probability of 0.5, the LAO angle was shifted by \ang{180} to capture a possible ambiguity in the projections. 

    \item \textit{Architecture:} To eliminate the need for the affine coupling blocks to learn the complete representation of the input images, a conditioning network was applied that transformed the two input images into an intermediate representation. The choice of the architecture of the conditioning network was inspired by~\cite{chee2018airnet}, where core elements of the registration network are blocks with convolutional layers followed by batch normalization, dropout layers \((p=0.2)\), and ReLU activations. In the first stage of the training, we pre-trained the conditioning network with a mean squared error loss to predict the pose parameters.

    The cINN consisted of three affine coupling blocks, each followed by a (fixed) random permutation. The subnetworks were implemented as fully-connected networks with a single hidden layer with 128 dimensions, dropout layers \(p=0.02\), and tanh activations. Soft clamping was applied with a constant of 1.9. 
    The cINN was trained with a maximum likelihood loss, batch size of 32, and noise and contrast augmentation for both CT volume and 2D projections. The model was trained for 3000 epochs with the Adam optimizer with a weight decay of \(10^{-4}\) and an initial learning rate of \(10^{-2}\). Every 200 epochs, the learning rate was reduced by a factor of two. During the training of the cINN, the conditioning network was further optimized.

    \item \textit{Mode Processing:} Upon test time, CT volume and 2D projection serve as conditioning input, and repeated sampling from the latent space (here: 1,028 samples) results in a full posterior over the five-dimensional parameter space. For mode detection, the DBSCAN clustering algorithm, as implemented in the scikit-learn library, was used. We fixed the parameter \(\varepsilon =0.19\) and varied the minimum sample size between 3 and 500 for hyperparameter optimization. For the localization criterion and the assignment strategy, we solely considered the LAO angle as this is the dimension with expected ambiguous solutions.
\end{itemize}

\subsection{Medical vision use case 2: Functional tissue parameter estimation}
\begin{itemize}

    \item \textit{Data Set:} The simulations were performed on 16 equidistant wavelengths between 700 and 850 nm. The optical Monte Carlo simulation was performed with \(5\cdot 10^8\) photons with a spatial resolution of 0.15625mm. The volumes were of dimension: 75mm (transducer dim) x 20mm (planar dim) x 20mm (height). The simulated 3D images were cropped, and additive and multiplicative Gaussian noise components were added to match the contrast of real photoacoustic images. Finally, the spectra of the tissue classes artery and vein were extracted, L\(_1\)-normalized, and used as input for our models.

    \item\textit{Architecture:} The original architecture of our baseline method (LSD) was adapted, resulting in a fully connected network with two hidden layers of size 256, dropout (\(p=0.5\)), and ReLU activations. 
    For the cINN, 20 coupling blocks and (fixed) random permutations were used. The subnetworks were implemented as fully connected networks with one hidden layer of size 1024, dropout (\(p=0.5\)), and ReLU activations. Soft clamping was applied with $\alpha=1.0$. As the coupling blocks require a minimum channel dimension of two due to the internal dimension splitting, a second dummy dimension with standard Gaussian noise was concatenated to the one-dimensional quantity of interest (oxygenation).  

    Both models were trained with a batch size of 1024 for 100 epochs. The AdamW optimizer was used with a learning rate of \(10^{-3}\) and weight decay of 0.01. After epochs 80 and 90, the learning rate was reduced by a factor of ten. 

    The UniDip clustering algorithm \citep{maurus2016skinny} was used with a statistical significance level of $\alpha=0.05$.
\end{itemize}

\bibliographystyle{elsarticle-harv} 
\bibliography{main.bib}

\end{document}